\pdfoutput=1

\documentclass[11pt]{article}

\usepackage{acl}

\usepackage{times}
\usepackage{latexsym}

\usepackage[T1]{fontenc}

\usepackage[utf8]{inputenc}

\usepackage{microtype}

\usepackage{graphicx}
\usepackage{stfloats}
\usepackage{natbib}
\usepackage{cancel}
\usepackage{bm}
\usepackage{amsmath}
\usepackage{array}
\usepackage{booktabs}
\usepackage{multirow}
\usepackage{makecell}
\usepackage{pifont}
\usepackage{color}
\usepackage{CJKutf8}

\newcommand{\PreserveBackslash}[1]{\let\temp=\\#1\let\\=\temp}
\newcolumntype{C}[1]{>{\PreserveBackslash\centering}p{#1}}
\newcolumntype{R}[1]{>{\PreserveBackslash\raggedleft}p{#1}}
\newcolumntype{L}[1]{>{\PreserveBackslash\raggedright}p{#1}}
\newcommand{\tabincell}[2]{\begin{tabular}{@{}#1@{}}#2\end{tabular}}
\newcommand{\chinese}[1]{\begin{CJK}{UTF8}{gkai}{}#1\end{CJK}}
\definecolor{myblue}{RGB}{91,155,213}
\definecolor{mygreen}{RGB}{146,208,80}

\title{Multitasking Framework for Unsupervised Simple Definition Generation}

\author{
	Cunliang Kong\textmd{\textsuperscript{134}}, 
	Yun Chen\textmd{\textsuperscript{2}},
	Hengyuan Zhang\textmd{\textsuperscript{1}},
	Liner Yang\textmd{\textsuperscript{134}\Thanks{~Correspondence to: Liner Yang.}},
	Erhong Yang\textmd{\textsuperscript{134}} \\
	\textsuperscript{1}School of Information Science, Beijing Language and Culture University \\
	\textsuperscript{2}School of Information Management \& Engineering, \\Shanghai University of Finance and Economics \\
	\textsuperscript{3}National Language Resources Monitoring and Research Center Print Media Branch, \\Beijing Language and Culture University \\
	\textsuperscript{4}Beijing Advanced Innovation Center for Language Resources, \\Beijing Language and Culture University
}

\begin{document}
\maketitle

\begin{abstract}
	The definition generation task can help language learners by providing explanations for unfamiliar words.
	This task has attracted much attention in recent years.
	We propose a novel task of Simple Definition Generation (SDG) to help language learners and low literacy readers.
	A significant challenge of this task is the lack of learner's dictionaries in many languages, and therefore the lack of data for supervised training.
	We explore this task and propose a multitasking framework \textbf{SimpDefiner} that only requires a standard dictionary with complex definitions and a corpus containing arbitrary simple texts.
	We disentangle the complexity factors from the text by carefully designing a parameter sharing scheme between two decoders.
	By jointly training these components, the framework can generate both complex and simple definitions simultaneously.
	We demonstrate that the framework can generate relevant, simple definitions for the target words through automatic and manual evaluations on English and Chinese datasets.
	Our method outperforms the baseline model by a 1.77 SARI score on the English dataset, and raises the proportion of the low level (HSK level 1-3) words in Chinese definitions by 3.87\% \footnote{Code can be found at \href{https://github.com/blcuicall/SimpDefiner}{https://github.com/blcuicall/\\SimpDefiner}.}.
\end{abstract}

\section{Introduction}
Helping language learners understand words in doubt is an important topic in the field of Intelligent Computer-Assisted Language Learning (ICALL) \citep{segler-2002-second, enayati-2020-impact, lolita-2020-impact}.
In recent years, researchers attempted to automatically generate definitions for words rather than formulating predefined word-definition inventories \citep{ishiwatari-2019-learning, yang-2020-incorp, huang-2021-definition}.
There are two reasons for this.
Firstly, it can be difficult for users to distinguish which sense is appropriate in the current context because of the cognitively inaccurate nature of discrete sense boundaries \citep{rosch-1975-family,kilgarriff-1997-don,tyler-2001-reconsidering}.
Secondly, the predefined inventories need to be updated manually by lexicographers, which is time-consuming and causes dictionaries to lag behind the ever-changing language usage.

\begin{figure}[t]
	\centering
	\includegraphics[width=\linewidth]{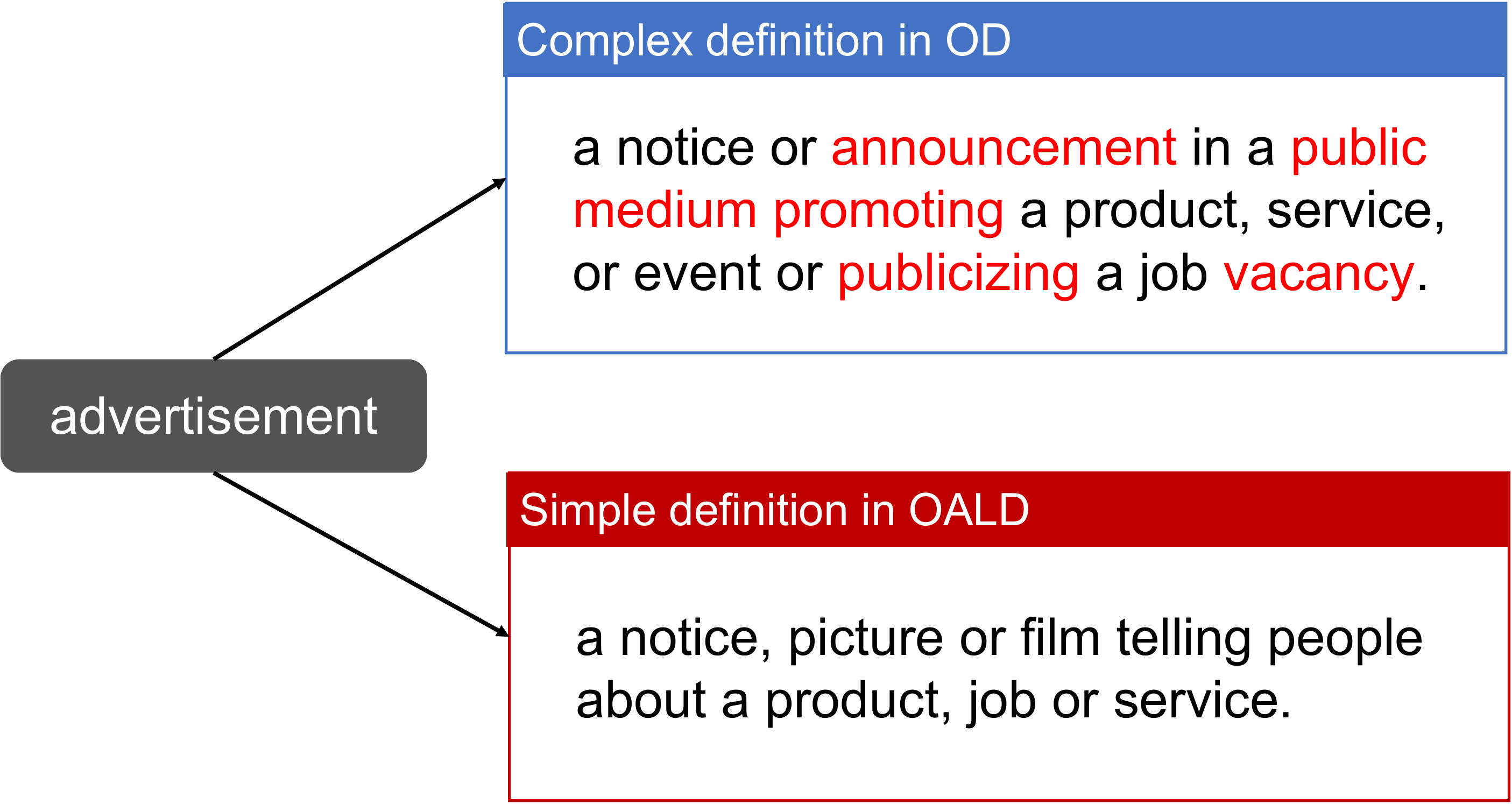}
	\caption{Different definitions for \textit{advertisement} in the Oxford Dictionary (OD) and Oxford Advanced Learner's Dictionary (OALD).}
	\label{fig:definitions}
\end{figure}

Different from previous work \citep{noraset-2017-definition, gadetsky-2018-conditional, mickus-2019-mark, kong-2020-toward} that focused only on how to generate definitions, we further propose a novel task of \textbf{S}imple \textbf{D}efinition \textbf{G}eneration (SDG).
Making the definitions easier to read and understand could benefit the language learners, low literacy readers, as well as helping people with aphasia or dyslexia.
For example, compared with the Oxford Dictionary (OD), the Oxford Advanced Learner's Dictionary (OALD) has simpler definitions, which are specifically designed for language learners.
As shown in Figure \ref{fig:definitions}, the definition of the word \textit{advertisement} in OALD does not contain difficult words or phrases such as \textit{announcement} and \textit{public medium}.

The goal of SDG task is to generate simple definitions for languages that lack learner's dictionary.
For example, Chinese as Second Language (CSL) learners do not have suitable dictionaries.
As \citet{zhang-2011-duiwai} pointed out, since the difficulty of definitions is not considered, the existing dictionary cannot meet CSL learner's needs.

The SDG task is challenging because it requires a model to learn from a standard dictionary containing complex definitions and then generate simple ones, and hence fully unsupervised.
A seemingly feasible solution is to generate definitions first and then simplify them, i.e., the generation-simplification pipeline.
However, the simplification task requires dataset with complex-simple sentence pairs, and such data is also difficult to find in languages other than English \cite{martin-2020-muss}.
Besides, the pipeline methods do not perform well due to accumulated errors (Section \ref{section:main-results}).

\begin{figure}[t]
	\centering
	\includegraphics[width=\linewidth]{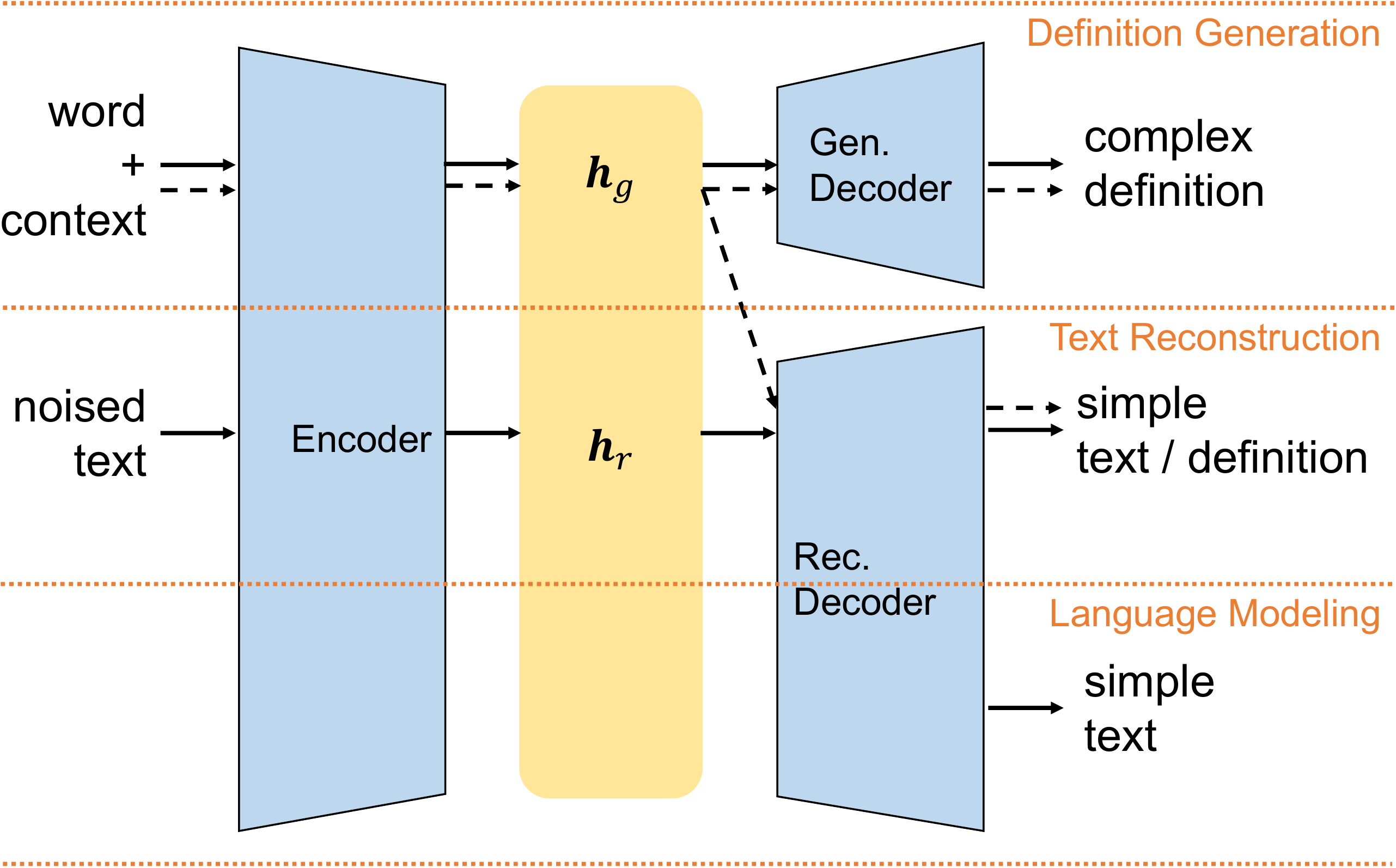}
	\caption{
		The SimpDefiner consists of three sub-tasks.
		Gen. means generation and Rec. means reconstruction.
		The solid black lines indicate the data-flow during training, and the dashed  black lines indicate the data-flow during inference.
	}
	\label{fig:arch}
\end{figure}

To solve this dilemma and bridge the gap between practical needs for simple definitions and current trivial definition generation systems, we present a novel method for the SDG task.
As illustrated in Figure \ref{fig:arch}, our method leverages a multitasking framework \textbf{SimpDefiner} to generate simple definitions by performing three sub-tasks at the same time, which are definition generation, text reconstruction, and language modeling tasks.
The framework consists of a fully shared encoder and two partially shared decoders.
We disentangle the complexity factors from the text by designing a parameter sharing scheme.
Particularly, we share parameters in Complexity-Dependent Layer Normalization and Complexity-Dependent Query Projection of the transformer architecture \citep{vaswani-2017-attention} to control the complexity (Section \ref{section:scheme}).
Through joint learning and sharing parameters between the decoders, the SimpDefiner is able to generate complex and simple definitions simultaneously.

Main contributions of our paper are listed below:
\begin{itemize}
	\item
	For the first time, we propose the task of SDG to generate simple definitions without supervised training data.
	
	\item
	We propose a multitasking framework SimpDefiner to tackle this task.
	Through joint training three sub-tasks, the framework can generate complex and simple definitions simultaneously.
	
	\item
	Both automatic and manual evaluations demonstrate the effectiveness of SimpDefiner.
	The framework outperforms the baseline model by 1.77 SARI score on the English test set.
	And the proportion of low level words (HSK level 1-3) in generated definitions raised by 3.87\% on the Chinese test set.
\end{itemize}

\section{Related Work}
\subsection{Definition Generation}
The definition generation task is first introduced by \citet{noraset-2017-definition}.
Although this task is proposed as a potentially useful tool for explainable AI, many subsequent works believe that it can assist language learning by giving definitions for words in the text \cite{ishiwatari-2019-learning, mickus-2019-mark,yang-2020-incorp}.

Various studies attempted to generate multiple different definitions for polysemous words.
\citet{gadetsky-2018-conditional} tackled this problem by computing the AdaGram vectors \cite{bartunov-2016-breaking} of input words, which are capable of learning different representations at desired semantic resolutions.
However, generating different definitions based on contexts, i.e., example sentences, became the mainstream method \cite{chang-2018-xsense, reid-2020-vcdm, li-2020-explicit, bevilacqua-2020-generationary}.
Among them, some studies used pre-trained language models to obtain contextualized embeddings.
\citet{reid-2020-vcdm} initialized encoders with BERT \cite{devlin-2019-bert} and employed variational inference for estimation and leveraged contextualized word embeddings for improved performance.
\citet{bevilacqua-2020-generationary} employed a novel span-based encoding scheme to fine-tune a pre-trained English encoder-decoder system to generate definitions.
\citet{huang-2021-definition} leveraged the T5 \cite{raffel-2019-exploring} model for this task and introduced a re-ranking mechanism to model specificity in definitions.

Our proposed SimpDefiner also takes the given word and context as input.
Differently, our main focus is to generate definitions with appropriate complexity to better help language learners.
Besides, our model is based on MASS \cite{song-2019-mass}, which is a pre-trained encoder-decoder model and is suitable for generation tasks.

\subsection{Sentence Simplification}
Researchers usually regard the sentence simplification task as a monolingual variant of machine translation (MT) \cite{wubben-2012-sentence}.
Benefiting from the advancement of neural machine translation, this task has also made great progress in recent years.

Lately, many works built upon the Seq2Seq MT model \cite{sutskever-2014-sequence} performed well.
First attempted by \citet{nisioi-2017-exploring}, the Seq2Seq models for this task are able to perform lexical simplification and content reduction simultaneously by training on complex-simple sentence pairs.
This method was inherited and improved by many subsequent works, such as combining with the reinforcement learning method by setting a simplification reward \cite{zhang-2017-sentence}, augmenting memory capacities \cite{vu-2018-sentence} or training with multitasking on entailment and paraphrase generation \cite{guo-2018-dynamic}.
\citet{martin-2019-access} proposed to prepend additional prompt tokens to source sentences at train time, which enables the end-users to condition the simplifications returned by the model on attributes like length, lexical complexity, and syntactic complexity.
This controllable simplification system (called ACCESS) and its improved version MUSS \cite{martin-2020-muss} achieved SOTA results on the Turk corpus in terms of the SARI metric \cite{xu-2016-optimizing}.

The generation-simplification pipeline methods are used as baselines of the SDG task, and we use both ACCESS and MUSS models for the simplification.
Unlike the baseline, the SimpDefiner can generate simple definitions directly, alleviating the accumulated errors.

\subsection{Unsupervised Style Transfer}
Style transfer aims to change the style attributes while preserving the content.
Our work is related to unsupervised style transfer by regarding the text complexity as one of the style attributes \cite{kawashima-2019-sentence}.

\citet{dumoulin-2017-alearned} demonstrated that the neural networks can capture the artistic style of a diversity of paintings.
The authors discovered that adjusting parameters in the layer normalization mechanism leads to different artistic styles.
This method permits users to transform images to arbitrary styles learned from individual paintings.
\citet{jin-2020-hooks} successfully applied this method to the task of headline generation, allowing the model to generate headlines of a specific style, such as humorous, romantic or click-baity, in an unsupervised manner.

By treating the task of simplification as a variant of style transfer, we borrow the insight of learning complexity-dependent parameters in the Layer Normalization mechanism.
Additionally, we introduce the language modeling task into SimpDefiner, which is to enhance the decoder and make it more sensitive to text complexity.

\section{Method}
We integrate three sub-tasks of definition generation, text reconstruction, and language modeling into the SimpDefiner.
This section first gives a formal definition of the SDG task, then introduces each sub-task, and finally the parameter sharing scheme.

\subsection{Task Formulation}
The SDG task is to generate a simple definition $\bm d^{sim}$ for a given word and context $(w^*, \bm c)$, where $\bm c = [w_1, \dots, w^*, \dots, w_n]$ is a sentence containing $w^*$.
This task is challenging because there is no corpus like $\{(w_i^*, \bm c_i, \bm d_i^{sim})\}_{i=1}^N$ and hence it is fully unsupervised.

The only data available in this work include a standard dictionary dataset $G = \{(w_i^*, \bm c_i, \bm d_i^{com})\}_{i=1}^N$ and a simple text corpus $Y=\{\bm y_i\}_{i=1}^M$.
Note that we use $\bm d^{com}$ for complex definitions and $\bm d^{sim}$ for simple ones.

\subsection{Multitasking Framework}
We design the three sub-tasks in the SimpDefiner to learn different abilities.
Cooperating with each other, the entire framework obtains the ability to compute the conditional probability $P(\bm d^{sim}|w^*, \bm c)$ of simple definitions in a zero-shot manner.

Specifically, the definition generation task aims to model the probability of a complex definition given the word and context $P(\bm d^{com} | w^*, \bm c)$ (Section \ref{section:dgt}).
And the text reconstruction task aims to model the probability of a simple sentence given the corrupted version $P(\bm y | \tilde{\bm{ y}})$ (Section \ref{section:trt}).
As we can see, neither task can directly get the $P(\bm d^{sim}|w^*, \bm c)$.
To solve the problem, we assume that complexity and semantic information are controlled by different parameters in the decoders, and we attempt to disentangle the complexity factors from the text through a carefully designed parameter sharing scheme.
In the inference stage, we obtain a simple definition by feeding the encoded hidden state into the reconstruction decoder as in Figure \ref{fig:arch}.
The detailed parameter sharing scheme is in Section \ref{section:scheme}.

Nevertheless, the complexity information may still be kept in some shared parameters, resulting in the reconstruction decoder fail to generate simple definitions occasionally.
Eliminating the complexity information in all shared parameters is obviously technically impossible.
Instead, we introduce the language modeling task (Section \ref{section:lmt}) to enhance the reconstruction decoder and make it more \textit{focused} on simple text generation.
The experiment results in Section \ref{section:results} confirm our assumption.



\subsubsection{Definition Generation Task} \label{section:dgt}
We follow the mainstream method \citep{yang-2020-incorp, kong-2020-toward, reid-2020-vcdm} to concatenate the word and context together with a special token [SEP] as $\mathbf x = (w^*;$ [SEP]$; \bm c)$.
The entire sequence is then fed into SimpDefiner, and the definition is obtained by the following language model:
\begin{equation}
\begin{aligned}
P(\bm d^{com}|\mathbf x; \bm \theta_g) = \prod_{t} P\left(\bm d^{com}_{t} | \bm d^{com}_{<t}, \mathbf x; \bm \theta_g \right),
\end{aligned}
\end{equation}
where $\bm d^{com}_{t}$ is the $t$-th token of the definition, and $\bm \theta_g$ is the set of parameters.
The model is optimized using the following loss function.
\begin{equation}
	\mathcal{L}_{gen}(\bm \theta_g) = -\sum_{(\mathbf x, \bm d^{com}) \in G} \log P(\bm d^{com} | \mathbf x; \bm \theta_g)
\end{equation}

\subsubsection{Text Reconstruction Task} \label{section:trt}
We corrupt each sentence in the corpus $Y$ by randomly deleting or blanking some words and shuffling the word orders.
And then we obtain a new corpus $\tilde Y = \{(\tilde{\bm y_i}, \bm y_i)\}_{i=1}^M$, and $\tilde{\bm{y}}$ is a corrupted version of $\bm y$.
We input $\tilde{\bm{y}}$ into SimpDefiner and obtain $\bm y$ by solving a self-supervised task of
\begin{equation} \label{eq:dae-lm}
P(\bm y|\tilde{\bm y}; \bm \theta_r) = \prod_{t} P\left(\bm y_{t} | \bm y_{<t}, \tilde{\bm y}; \bm \theta_r \right),
\end{equation}
where $\bm y_{t}$ is the $t$-th token of the sentence, and $\bm \theta_r$ is a set of parameters.
The loss function of this task is as follows:
\begin{equation}
\mathcal{L}_{rec}(\bm \theta_r) = -\sum_{(\bm y, \tilde{\bm y}) \in \tilde{Y}} \log P(\bm y | \tilde{\bm{y}}; \bm \theta_r).
\end{equation}

\subsubsection{Language Modeling Task} \label{section:lmt}
This task facilitates zero-shot generation of $P(\bm d^{sim}|\mathbf x)$ by jointly training the reconstruction decoder as a language model.
Once the model captures correct complexity that guides the model to generate the desired simple texts, it's more likely for the model to ignore the wrongly shared complexity information.
Similar to Eq. \ref{eq:dae-lm}, we have:
\begin{equation}
P(\bm y|\bm \theta_l) = \prod_{t} P\left(\bm y_{t} | \bm y_{<t}; \bm \theta_l \right).
\end{equation}
It is equivalent to masking the encoder out and ignoring the attention modules between the encoder and reconstruction decoder. The model is optimized by the following loss function:
\begin{equation}
\mathcal{L}_{lm}(\bm \theta_l) = -\sum_{\bm y \in Y} \log P(\bm y | \bm \theta_l).
\end{equation}

Finally, we train the entire SimpDefiner by jointly minimizing the weighted sum of all above mentioned loss functions.
And the overall loss function is calculated as:
\begin{equation} \label{eq:loss}
\mathcal{L} = \lambda_{\alpha} \mathcal{L}_{gen} + \lambda_{\beta} \mathcal{L}_{rec} + \lambda_{\gamma} \mathcal{L}_{lm},
\end{equation}
where $\lambda_{\alpha}$, $\lambda_{\beta}$, $\lambda_{\gamma}$ are hyper-parameters.

\begin{figure}[!t]
	\centering
	\includegraphics[width=0.65\linewidth]{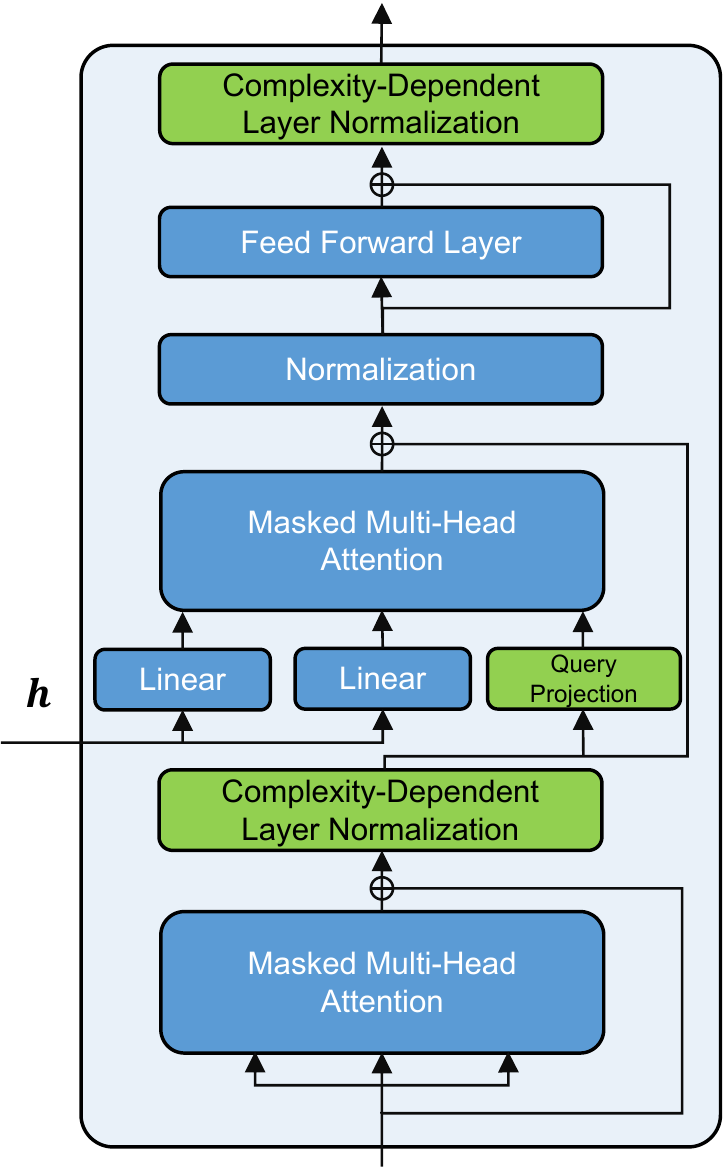}
	\caption{
		The parameter-sharing scheme between decoders.
		Parameters in \textcolor{myblue}{blue} layers are shared, and those in \textcolor{mygreen}{green} are not.
	}
	\label{fig:param}
\end{figure}

\subsection{Parameter-Sharing Scheme} \label{section:scheme}
%

For parameters in the decoders, we divided them into two parts, which are complexity-independent and complexity-dependent parameters.
The former ones are shared between decoders, and the latter ones are not, as illustrated in Figure \ref{fig:param}.

We now introduce the complexity-dependent layers, namely Complexity-Dependent Layer Normalization and Complexity-Dependent Query Projection.

\paragraph{Complexity-Dependent Layer Normalization}
Previous works \citep{dumoulin-2017-alearned, jin-2020-hooks} demonstrated that the layer normalization is related to the style of the target texts.
We further argue that as an attribute of style, the complexity can be retained by independent layer normalization.
Thus, we make the scaling and shifting parameters for layer normalization not shared in both decoders.
This approach is to transform a layer activation $\bm x$ into a complexity-specific normalized activation $\bm z$ as:
\begin{equation}
\bm z = \gamma_c (\frac{\bm x - \mu}{\sigma}) - \beta_c,
\end{equation}
where $\mu$, $\sigma$ are the mean and standard deviation of the batch of $\bm x$, and $\gamma_c$, $\beta_c$ are learnable parameters specific to complexity $c$.
Note that $c$ is a binary variable indicating different decoders.
This mechanism is used in all decoder layers.

\paragraph{Complexity-Dependent Query Projection}
The decoder layers extract information from encoded hidden states through cross-attention mechanism.
We believe that the required information may vary for different complexity.
Therefore, the parameters of the linear mapping used for the query transformation in the cross-attention are not shared among decoders.
This calculation is as follows:
\begin{equation}
\bm Q = \bm{\hat{Q}} \cdot \bm{W}_c^q,
\end{equation}
where $\bm W_c^q$ is the query transformation matrix specific to complexity $c$.
The obtained query vector $\bm Q$ is then fed into the cross-attention mechanism.
By using this approach, the model can obtain different information from the encoded hidden states for different complexities.
\begin{table*}[!t]
	\centering
	\footnotesize
	\begin{tabular}{lC{1.1cm}C{1.1cm}C{1.1cm}rC{1.1cm}rC{1.1cm}C{1.1cm}C{1.1cm}}
		\hline
		& \multicolumn{3}{c}{OD} & & \makecell[c]{OALD} & & \multicolumn{3}{c}{CWN} \\ 
		\cmidrule{2-4} \cmidrule{6-6} \cmidrule{8-10}
		& \makecell[c]{Train} & \makecell[c]{Valid} & \makecell[c]{Test} & & \makecell[c]{Test} & & \makecell[c]{Train} & \makecell[c]{Valid} & \makecell[c]{Test} \\
		\hline
		Words & 33,128 & 8,867 & 3,881 & & 3,881 & & 6,574 & 823 & 823 \\
		Entries & 97,855 & 12,232 & 5,111 & & 5,111 & & 67,861 & 8,082 & 8,599 \\
		Context Length & 17.74 & 17.80 & 16.24 & & 16.24 & & 34.49 & 34.73 & 34.06 \\
		Def. Length & 11.02 & 10.99 & 10.03 & & 12.74 & & 14.76 & 14.60 & 14.72 \\
		\hline
	\end{tabular}
	\caption{Statistics of the OD (English) dataset, OALD (English) test set, and CWN (Chinese) dataset. The rows are number of words and entries, and the average length of contexts and definitions.}
	\label{table:main-data}
\end{table*}


\begin{table}[t]
	\centering
	\footnotesize
	\begin{tabular}{L{1.5cm}R{1.5cm}R{1.5cm}R{1.5cm}}
		\hline
		& Sents & Tokens & Avg. Len \\
		\hline
		English & 32,395 & 392,625 & 12.12 \\
		Chinese & 58,867 & 860,761 & 14.62 \\
		\hline
	\end{tabular}
	\caption{Statistics of simple text corpora. The columns are number of sentences and tokens, and the average length of sentences.}
	\label{table:aux-data}
\end{table}

\section{Datasets}
We evaluate the proposed multitasking framework on both English and Chinese datasets.
Each language has a definition generation dataset and a simple text corpus.

\subsection{English Dataset}
The English datasets are constructed from the Oxford Dictionary (OD) and Oxford Advanced Learner's Dictionary (OALD).
Since the OALD is for language learners, it has much simpler definitions than OD.
Therefore, we use the OD for the definition generation training, and use the OALD for validation of simple definition generation.
Note that the words used for testing are excluded from the training and validation sets.

For the definition generation dataset, we directly use the OD dataset published by \citet{gadetsky-2018-conditional}.
The training set has 33,128 words and 97,855 entries.
Each entry consists of a triplet of $(w^*, \bm c, \bm d^{com})$.
For testing, we align the words and context in OD with the definitions in OALD through manual annotation.
The annotated test set includes 3,881 words and 5,111 entries, which is used for automatic evaluation in experiments.
Each entry in the test set has both golden complex and simple definitions from OD and OALD, respectively.
Detailed statistics are listed in Table \ref{table:main-data}.

We extract the OALD definitions that are not in the test set for constructing the simple text corpus.
This corpus has 32,395 sentences with an average length of 12.12.
We list more statistics in Table \ref{table:aux-data}.

During training, the definition generation dataset and the simple text corpus are randomly sampled as mini-batches respectively.
And there is no alignment between the two mini-batches at each step.

\subsection{Chinese Dataset}
For the definition generation dataset, we use the Chinese WordNet (CWN) \citep{huang-2010-chinese}, which is a semantic lexicon aiming to provide a knowledge base of sense distinction.\footnote{Chinese WordNet: \url{http://lope.linguistics.ntu.edu.tw/cwn2}}
We use the corresponding words, contexts, and definitions in CWN for the definition generation task.
We split the entire dataset into training, validation, and test sets roughly according to the ratio of 8:1:1.
The training set contains 6,574 words and 67,861 entries.
Statistics are listed in Table \ref{table:main-data}.

For the simple text corpus, we extract 58,867 sentences from a number of primary level Chinese as Second Language textbooks, with an average sentence length of 14.62.

Since no suitable dictionary can be used for evaluation, there are no golden simple definitions in Chinese Dataset.
In the experiments, we count the difficulty level of words in definitions to estimate if they are simple.
We also organize a manual evaluation to score the accuracy and simplicity of definitions.

\section{Experiments}
This section presents the experimental settings and evaluation methods.

\subsection{Settings}

\paragraph{Baselines}
We compare the SimpDefiner with generation-simplification pipelines.
We first employ LOG-CaD \citep{ishiwatari-2019-learning} and MASS \citep{song-2019-mass} models to generate definitions, and then employ ACCESS \citep{martin-2019-access} and MUSS \citep{martin-2020-muss} models to simplify them.
Thus, we have four different pipeline baselines.
Since these models are not available in Chinese, we only apply these pipelines to English datasets.
For the Chinese SDG task, we specially pretrained a MASS-ZH model from scratch using the \textit{Chinese Gigaword Fifth Edition}\footnote{https://catalog.ldc.upenn.edu/LDC2011T13} corpus.
Note that we set the learning rate to 3e-4, warmup steps to 500 when fine-tuning both MASS and MASS-ZH.
%

\paragraph{SimpDefiner}
We use the parameters in the MASS model to initialize the encoder and two decoders in SimpDefiner.
For the sentence corruption in the text reconstruction task, we randomly delete or blank words with a uniform probability of 0.2, and randomly shuffle the order of words within 5 tokens.
For the language modeling task, we set the input representations to $\bm{0}$ and use the simplified text as the target output.
We tune the $\lambda$ parameters in Eq. \ref{eq:loss} on the validation set and adopt the same hyper-parameters as the baseline for comparison.
We set 5 different random seeds as and report the average result of multiple runs.
Each run takes 7.68 GPU hours on 4 GeFource RTX 2080 Ti GPUs.


\subsection{Evaluation}
Evaluation of the generated definitions mainly focuses on two aspects, i.e., accuracy and simplicity.
We perform both automatic and manual evaluations for each aspect.

We first introduce these automatic metrics, and then the manual evaluation method.

\paragraph{BLEU}
Previous definition generation studies \cite{noraset-2017-definition, yang-2020-incorp, kong-2020-toward} used the BLEU \cite{papineni-2002-bleu} score to measure the closeness of generated results to the standard answers, and to evaluate the accuracy of results.
Since the English test set is manually annotated,  we calculate the BLEU score of both complex and simple definitions, respectively.

\paragraph{Semantic Similarity}
In addition to the BLEU score, we use the sentence-transformers toolkit \cite{reimers-2020-multilingual} to convert the generated definitions and references into sentence vectors, and calculate cosine similarity between them.

\paragraph{SARI}
SARI \cite{xu-2016-optimizing} is a lexical simplicity metric that measures how good are the words added, deleted and kept by a simplification model.
This metric compares the model output to simplification references and the original sentence.
We use the SARI implementation in the EASSE toolkit\footnote{\url{https://github.com/feralvam/easse}}.


\paragraph{HSK Level}
HSK, namely Chinese Proficiency Test, is set up to test the proficiency of non-native speakers\footnote{\url{http://www.chinesetest.cn}}.
It has nine levels, from easy to hard, and each level corresponds to a vocabulary.
We count the proportion of words at levels 1-3 and 7+ in the generated definitions.
The higher the proportion of words in levels 1-3 (7+), the easier (more challenging) the definitions are understood.

\paragraph{Manual Evaluation}
We randomly select 200 words and contexts from the Chinese test set and let the MASS and SimpDefiner generate definitions for them one by one.
We mix the two generated definitions and the golden complex definition and then ask three native-speaker annotators to score them.
Specifically, each annotator evaluates the definitions on two criteria of accuracy and simplicity.
Both criteria have a range of 1-3.
For accuracy, the annotators are asked to evaluate how semantically relevant the definitions are to the word.
For simplicity, the annotators are asked to evaluate how simple the definitions are.
After collecting evaluation results, we average the scores as final score.

\section{Results and Analysis} \label{section:results}

\begin{table}[!t]
	\centering
	\footnotesize
	\begin{tabular}{L{1.8cm}|C{0.6cm}C{0.6cm}|C{0.6cm}C{0.6cm}|C{0.6cm}}
		\hline
		& \multicolumn{2}{c|}{Complex} & \multicolumn{2}{c|}{Simple} & \multirow{2}{*}{SARI} \\
		\cline{2-5}
		& BLEU & SSim & BLEU & SSim & \\
		\hline
		LOG-CaD & 19.04 & 40.32 & -- & -- & -- \\
		~~+ ACCESS & -- & -- & 12.32 & 32.63 & 38.02 \\
		~~+ MUSS & -- & -- & 11.74 & 27.66 & 36.53 \\
		\hline
		MASS & 24.00 & 52.78 & -- & -- & -- \\
		~~+ ACCESS & -- & -- & 12.95 & 38.53 & 38.59 \\
		~~+ MUSS & -- & -- & 12.58 & 37.49 & 38.48 \\
		\hline
		SimpDefiner & \textbf{24.17} & \textbf{53.87} & \textbf{15.05} & \textbf{46.99} & \textbf{40.36} \\
		\hline
	\end{tabular}
	\caption{Main results on the English test set. LOG-CaD \citep{ishiwatari-2019-learning} is a definition generation model.}
	\label{table:result-en-auto}
\end{table}

\begin{table}[!t]
	\centering
	\footnotesize
	\begin{tabular}{L{1.5cm}|R{1.5cm}R{1.5cm}}
		\hline
		& L1-3 (\%) & L7+ (\%) \\
		\hline
		MASS & 44.16 & 37.05 \\
		SimpDefiner & \textbf{48.03} & \textbf{36.59} \\
		\hline
	\end{tabular}
	\caption{Main results on the Chinese test set.}
	\label{table:result-zh-auto}
\end{table}

\begin{table}[!t]
	\centering
	\footnotesize
	\begin{tabular}{l|l|R{0.7cm}R{0.7cm}R{0.7cm}R{0.7cm}}
		\hline
		& & \#1 & \#2 & \#3 & Avg. \\
		\hline
		\multirow{3}{*}{Acc.} & Golden & 3.00 & 2.93 & 2.98 & 2.97 \\
		& MASS & 1.26 & 1.30 & 1.38 & 1.31 \\
		& SimpDefiner & 1.48 & 1.47 & 1.59 & \textbf{1.51} \\
		\hline
		\multirow{3}{*}{Sim.} & Golden & 2.04 & 2.06 & 2.11 & 2.07 \\
		& MASS & 1.92 & 2.03 & 1.89 & 1.95 \\
		& SimpDefiner & 2.14 & 2.04 & 2.21 & \textbf{2.13} \\
		\hline
	\end{tabular}
	\caption{Manual evaluation results on the Chinese test set. Accuracy and simplicity scores are listed in the table. The last column are averaged scores among all three annotators. }
	\label{table:result-manual}
\end{table}

\subsection{Main Results} \label{section:main-results}
Table \ref{table:result-en-auto} and Table \ref{table:result-zh-auto} present the experiment results on the English and Chinese test sets respectively.
Results show that our proposed SimpDefiner significantly outperforms baseline methods of generation-simplification pipelines on both English and Chinese datasets.

For English results, the performance of simple definition generation improves 2.1 and 8.46 on the BLEU and SemSim metrics respectively, and improves 1.77 on the SARI metric.
This indicates that both accuracy and simplicity are effectively improved comparing with the baseline.
We also observe that complex definition generation also improves by 0.17 on BLEU and 1.09 on SemSim.
This shows that SimpDefiner improves the ability to generate both complex and simple definitions.

For Chinese results, we compute the HSK Level metric on generated simple definitions.
We can see that the proportion of low-level (HSK level 1-3) words increases by 3.87\%, and that of high-level (HSK level 7+) words decreases by 0.46\%.
The lexical complexity of the SimpDefiner generated definitions are significantly reduced.

Besides, we also conduct a manual evaluation on the Chinese test set, and the results are listed in Table \ref{table:result-manual}.
From the averaged scores, we observe that SimpDefiner outperforms MASS by 0.2 in terms of accuracy (more accurate) and 0.18 in terms of simplicity (more straightforward).
On the accuracy score, all three annotators agree that SimpDefiner has higher accuracy than MASS, which shows the superiority of our framework.
As expected, the golden definitions have the highest accuracy in the table, far exceeding the definitions generated by the two models.
We believe this is caused by insufficient knowledge in the model, and this can be solved by using larger pretrained models, such as BART \cite{lewis-2019-bart}.
On the simplicity score, three annotators agree that SimpDefiner generates simpler definitions than MASS, and two of three annotators think SimpDefiner generates simpler definitions than the golden ones.


\begin{table}[t]
	\footnotesize
	\centering
	\begin{tabular}{R{0.2cm}|L{1.45cm}|R{0.6cm}R{0.6cm}|R{0.6cm}R{0.6cm}|R{0.6cm}}
		\hline
		\multirow{2}{*}{ID} & \multirow{2}{*}{Model} & \multicolumn{2}{c|}{Complex} & \multicolumn{2}{c|}{Simple} & \multirow{2}{*}{SARI} \\
		\cline{3-6}
		& & BLEU & SSim & BLEU & SSim &  \\
		\hline
		\ding{172} & SimpDefiner & 24.31 & 53.60 & \textbf{15.24} & \textbf{47.05} & \textbf{40.19} \\
		\ding{173} & \ding{172} -- LM & 23.83 & 53.04 & 14.82 & 45.74 & 39.63 \\
		\ding{174} & \ding{173} -- TR & \textbf{25.02} & 53.80 & 13.66 & 44.01 & 38.58 \\
		\hline
		\ding{175} & \ding{172} -- LN & 24.45 & 53.76 & 13.87 & 44.66 & 38.61 \\
		\ding{176} & \ding{172} -- QP & 23.40 & 52.95 & 14.61 & 45.57 & 39.87 \\
		\ding{177} & \ding{175} -- QP & 24.80 & \textbf{53.95} & 13.90 & 44.77 & 38.52 \\
		\hline
	\end{tabular}
	\caption{Ablation study on the English test set. LM: the language modeling task. TR: the text reconstruction task. LN: complexity-dependent layer normalization. QP: complexity-dependent query projection.}
	\label{table:result-en-ablation}
\end{table}


\subsection{Ablation Study} \label{section:ablation}
We conduct ablation experiment to demonstrate the effectiveness of SimpDefiner components and the parameter sharing scheme.
For the language modeling (LM) and text reconstruction (TR) tasks, we ablate them by setting their weights to 0.
For the layer normalization (LN) and query projection (QP) as parameter-shared layers, we ablate them by sharing their parameters between models.
We illustrate the experiment results in Table \ref{table:result-en-ablation}.

In general, ablating any of the components or parameter-shared layers reduces the performance in terms of simple definitions, which indicates that the SimpDefiner benefits from both components and parameter sharing scheme.
We also observe that the performance of ablation experiments have slight disturbance on complex definitions.
But since we pay more attention to the performance on simple definitions, we argue that the benefits of SimpDefiner far outweigh the losses.

\begin{table}[t]
	\centering
	\footnotesize
	\begin{tabular}{L{1.5cm}|R{0.65cm}R{0.65cm}|R{0.65cm}R{0.65cm}|R{0.65cm}}
		\hline
		\multirow{2}{*}{($\lambda_\alpha$, $\lambda_\beta$, $\lambda_\gamma$)}& \multicolumn{2}{c|}{Complex} & \multicolumn{2}{c|}{Simple} & \multirow{2}{*}{SARI} \\
		\cline{2-5}
		& BLEU & SSim & BLEU & SSim & \\
		\hline
		(0.8,0.1,0.1) & \textbf{24.31} & \textbf{53.60} & \textbf{15.24} & \textbf{47.05} & 40.19 \\
		(0.6,0.2,0.2) & 23.27 & 53.19 & 15.01 & 46.85 & 40.49 \\
		(0.4,0.3,0.3) & 21.92 & 51.82 & 15.11 & 46.54 & \textbf{40.74} \\
		\hline
	\end{tabular}
	\caption{Different hyper-parameter settings on the English test set.}
	\label{table:result-en-weights}
\end{table}

\subsection{Analysis on Hyper-Parameters}
Furthermore, we conduct additional experiments on the English dataset to study how hyper-parameters affect the performance.
By setting different $\lambda$ to each model, we observe the relationship between the performance and these weights.

The experiment results are listed in Table \ref{table:result-en-weights}.
From the table, we observe the inconsistency between metrics.
As the definition generation task weight declines, the BLEU and SemSim metrics are generally declining, but the SARI metric is increasing.
Since the BLEU and SemSim measure the accuracy and the SARI measures simplicity, we consider this phenomenon as a seesaw between the two attributes of accuracy and simplicity.
The balance between them can be achieved by conditioning the hyper-parameters.

\begin{table}[!t]
	\centering
	\footnotesize
	\begin{tabular}{L{1.5cm}|L{5.5cm}}
		\hline
		Word & commander \\
		\hline
		Context & \tabincell{l}{Military commanders have warned coalition\\ troops in the south.} \\
		\hline
		Golden & \tabincell{l}{A person who is in charge of sth, especially\\ an officer in charge of a particular group of\\ soldiers or a military operation.} \\
		\hline
		Baseline & \tabincell{l}{An officer of the highest rank is a country\\in a country.} \\
		\hline
		\textbf{SimpDefiner} & The head of a military force. \\
		\hline
		\hline
		Word & \tabincell{l}{\chinese{督促} (supervise and urge)} \\
		\hline
		Context & \tabincell{l}{\chinese{我很感谢他的支持、鼓励与督促。} \\ I appreciate his support, encouragement\\and supervision.} \\
		\hline
		Golden & \tabincell{l}{\chinese{监督他人并促使后述事件发生。}\\Supervise others and promote the \\ occurrence of the following events.} \\
		\hline
		Baseline & \tabincell{l}{\chinese{以后述对象为凭借进行特定事件。}\\Sb. is used as a reference for specific events.} \\
		\hline
		\textbf{SimpDefiner} & \tabincell{l}{\chinese{要求后述对象赶快行动。}\\Ask sb. to act quickly.} \\
		\hline
	\end{tabular}
	\caption{Cases of generated simple definitions.}
	\label{table:case}
\end{table}

\subsection{Case Study}
Table \ref{table:case} shows two generation cases from English and Chinese test set respectively.
In both cases, the golden definition is a long sentence with quite complicated syntax.
The baseline generated definitions contains difficult words and often wrongly defines the given word.
In the English case, the word \textit{commander} is defined by the baseline as \textit{an officer of the highest rank in a country}, which is incorrect in most cases.
In the Chinese case, the baseline generated definition contains difficult words like \chinese{凭借} \textit{(reference)} and \chinese{特定事件} \textit{(specific events)}.
On the other hand, the SimpDefiner generates simple and accurate definitions in both cases.

\section{Conclusion}
In this work, we propose the SDG task, a novel task of generating simplified definitions in a zero-shot manner.
To this end, we leverage a multitasking framework SimpDefiner to tackle this task.
We introduce a text reconstruction task to the framework to control the text complexity, and a language modeling task to enhance the decoder.
For evaluation, we construct a novel test set in English by manually aligning the two dictionaries of OD and OALD.
The automatic and manual evaluations indicate that the our proposed framework can generate more accurate and more straightforward definitions than other models and the generation-simplification pipelines.
In the future, we will try to combine the current method with prompt learning methods, aiming to let users condition the complexity of generated definitions.

\section*{Acknowledgements}
This work was supported by the funds of Beijing Advanced Innovation Center for Language Resources (No. TYZ19005), Research Project of the National Language Commission (No. ZDI135-131) and National Natural Science Foundation of China (No. 62106138, No. 61872402).
We would like to thank Xiaowan Wang, Chenhui Xie, and Junhui Zhu for their manual evaluation and all anonymous reviewers for their valuable comments and suggestions on this work.

\bibliographystyle{acl_natbib}
\bibliography{custom}

\end{document}